\documentclass[conference]{IEEEtran}
\IEEEoverridecommandlockouts
\usepackage{hyperref}
\usepackage{url}

\usepackage{algorithm}
\usepackage{algpseudocode}
\usepackage{amsfonts}
\usepackage{graphicx}
\usepackage{caption}

\usepackage{multirow}
\usepackage{booktabs}
\usepackage{arydshln}

\usepackage{lscape}

\usepackage{subcaption}

\usepackage{float}

\usepackage{cite}
\usepackage{amsmath,amssymb,amsfonts}
\usepackage{graphicx}
\usepackage{textcomp}
\usepackage{xcolor}
\def\BibTeX{{\rm B\kern-.05em{\sc i\kern-.025em b}\kern-.08em
    T\kern-.1667em\lower.7ex\hbox{E}\kern-.125emX}}
\begin{document}

\title{Model Attribution in LLM-Generated Disinformation: A Domain Generalization Approach with Supervised Contrastive Learning
\thanks{The corresponding author.}
}

\author{\IEEEauthorblockN{Alimohammad Beigi$^{*}$}
\IEEEauthorblockA{\textit{School of Computing and AI } \\
\textit{Arizona State University}\\
Tempe, AZ \\
abeigi@asu.edu}
\and
\IEEEauthorblockN{Zhen Tan}
\IEEEauthorblockA{\textit{ School of Computing and AI } \\
\textit{Arizona State University}\\
Tempe, AZ \\
ztan36@asu.edu}
\and
\IEEEauthorblockN{Nivedh Mudiam}
\IEEEauthorblockA{\textit{School of Computing and AI } \\
\textit{Arizona State University}\\
Tempe, AZ \\
nmudiam@asu.edu}
\and
\IEEEauthorblockN{Canyu Chen}
\IEEEauthorblockA{\textit{Department of Computer Science} \\
\textit{Illinois Institute of Technology}\\
Chicago, IL \\
cchen151@hawk.iit.edu}
\and
\IEEEauthorblockN{Kai Shu}
\IEEEauthorblockA{\textit{Department of Computer Science} \\
\textit{Emory University}\\
Atlanta, GA \\
kai.shu@emory.edu}
\and
\IEEEauthorblockN{Huan Liu}
\IEEEauthorblockA{\textit{School of Computing and AI} \\
\textit{Arizona State University}\\
Tempe, AZ \\
huanliu@asu.edu}
}

\maketitle

\begin{abstract}
Model attribution for LLM-generated disinformation poses a significant challenge in understanding its origins and mitigating its spread. 
This task is especially challenging because modern large language models (LLMs) produce disinformation with human-like quality. Additionally, the diversity in prompting methods used to generate disinformation complicates accurate source attribution. These methods introduce domain-specific features that can mask the fundamental characteristics of the models. 
In this paper, we introduce the concept of model attribution as a domain generalization problem, where each prompting method represents a unique domain. We argue that an effective attribution model must be invariant to these domain-specific features. It should also be proficient in identifying the originating models across all scenarios, reflecting real-world detection challenges. To address this, we introduce a novel approach based on Supervised Contrastive Learning. This method is designed to enhance the model's robustness to variations in prompts and focuses on distinguishing between different source LLMs. We evaluate our model through rigorous experiments involving three common prompting methods: ``open-ended'', ``rewriting'', and ``paraphrasing'', and three advanced LLMs: ``llama 2'', ``chatgpt'', and ``vicuna''. Our results demonstrate the effectiveness of our approach in model attribution tasks, achieving state-of-the-art performance across diverse and unseen datasets.
\end{abstract}


\section{Introduction}
%Disinformation is harmfull and detection is important
The rapid advancement of artificial intelligence has compromised the accuracy of information. This results in the rise of LLM-generated disinformation, which significantly challenges societal trust and information integrity\cite{chen2023can,jiang2024catching,jiang2024media}. Disinformation refers to false information that is intentionally created and spread with the purpose of deceiving or misleading audiences\cite{fallis2015disinformation, wu2019misinformation}. A significant challenge involves tracing disinformation back to its originating models, especially when these models are capable of generating high-quality textual content that can compete with human authorship\cite{jiang2024disinformation, huang2024can}.

\begin{figure}[t]
\vspace{-2mm}
	\centering
	\includegraphics[width=0.95\columnwidth]{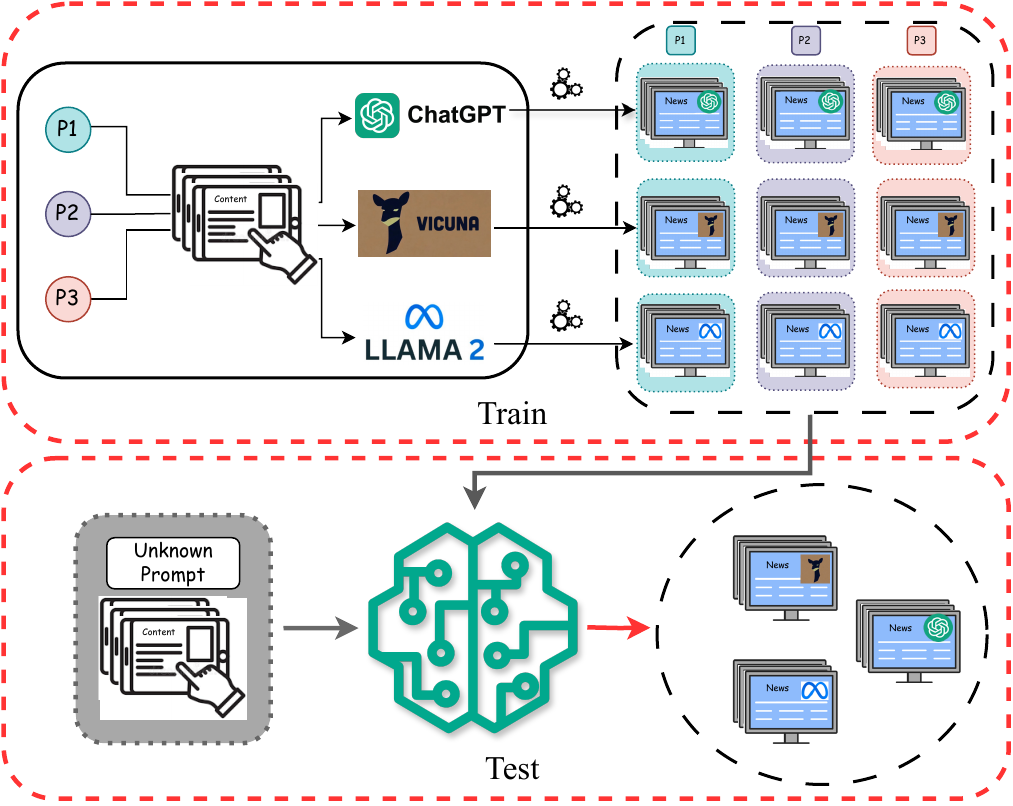}
    % \vspace{-4mm}
	\caption{A schematic diagram of model attribution in LLM-generated disinformation.}
	\label{fig:model}
	\vspace{-3mm}
\end{figure}

%LLMs bring new challenges to this area
\noindent The task of attributing LLM-generated disinformation to its source model is complicated by two primary factors. Firstly, modern Large Language Models (LLMs) like GPT-4 and LLaMA-2 can produce text that often rivals human-written content in quality\cite{kumarage2024survey}. This high quality makes it challenging to clearly identify and attribute their outputs. Secondly, and more significantly, the variety of prompting methods, spanning from open-ended to specific instructions such as rewriting or paraphrasing, adds further complexity\cite{chen2023can}. These methods influence the stylistic and substantive features of the generated content. They complicate the task of model attribution through textual analysis alone\cite{kumarage2024survey}. Additionally, these models enable the rapid creation and dissemination of large volumes of disinformation, which can easily surpass the capabilities of traditional detection systems.

%Our first/major contribution, formulate DG
\noindent Recognizing these challenges, we formulate the LLM attribution as a domain generalization problem. Domain generalization is a machine learning paradigm where the goal is to train a model that can generalize well to new, unseen domains by learning domain-invariant features from multiple source domains. Unlike domain adaptation, which assumes access to target domain data during training, domain generalization does not have any information about the target domain beforehand. This is particularly relevant for our task because it mirrors the unpredictability and variability of real-world scenarios. Each prompting method introduces domain-specific features that an ideal attribution model should be robust against. For instance, different prompting techniques might generate disinformation with varying linguistic styles, syntactic structures, or contextual details. A robust attribution model must identify the underlying patterns that are consistent across these variations, effectively generalizing across domains without being misled by superficial differences. By formulating the problem this way, we ensure that our model is not overfitted to specific types of prompts or methods used during the training phase. Instead, it learns to recognize the core attributes of disinformation regardless of how it was generated.

\noindent To address this problem, we propose a Supervised Contrastive Learning (SCL) for text classification approach that considers each prompting method as a unique domain with specific features (Figure~\ref{fig:model}). SCL optimizes the model by pulling together representations of instances from the same class and pushing apart those from different classes within the embedding space. This method is essential for extracting domain-invariant features, thereby refining the decision boundaries of our classifier and enhancing its accuracy and generalizability across varied disinformation scenarios. The key innovation of our methodology is creating a model invariant to these domain-specific features while focusing on the underlying characteristics that define a specific LLM. This invariance is crucial for ensuring that the attribution model performs consistently across different types of prompts. 

\noindent In this work, we consider three typical prompting methods - ``open-ended'', ``rewriting'', and ``paraphrasing'' - and evaluate them on three state-of-the-art LLMs: LLaMA-2, ChatGPT, and Vicuna. The results show the effectiveness of our model in discerning the outputs of these LLMs across diverse prompting conditions. Furthermore, we extend our investigation to case studies involving ChatGPT-4, exploring its in-context learning capabilities for model attribution. These studies show the potential of ChatGPT-4 to recognize and adapt to the writing styles characteristic of specific LLMs. However, they also reveal the inherent limitations of this strategy, such as its success being dependent on the quality and relevance of the training examples provided during the learning process.

%Summary
The key contributions of this work are outlined as follows:
\begin{itemize}
    \item We introduce the LLM-generated disinformation attribution as a domain generalization problem, treating each prompting method as a distinct domain.
    \item We develop a new method that enhances model robustness against varied prompting styles, focusing on identifying source LLMs based on their textual signatures.
    \item We validate the model across diverse prompting methods: ``open-ended'', ``rewriting'', and ``paraphrasing'' and three LLMs: ChatGPT, LLaMA-2, and Vicuna, demonstrating robust performance.
    \item We explore ChatGPT-4 in-context learning for source detection, assessing its strengths and limitations in practical scenarios.
\end{itemize}

\section{Related Work}

\subsection{Domain Adaptation}
Domain adaptation (DA) aims to minimize the distributional discrepancy between the source and target domains, enabling a model trained on a labeled source domain to perform well on a different, unlabeled target domain. Traditional approaches often utilize kernel methods such as Maximum Mean Discrepancy (MMD) and spectral feature alignment to achieve feature alignment across domains\cite{arbel2019maximum, pan2010spectral}. MMD measures the difference between two probability distributions based on their means in a reproducing kernel Hilbert space, helping to reduce the domain gap\cite{borgwardt2006integrating}. Spectral feature alignment, on the other hand, aligns the source and target domain features by using a shared latent space\cite{pan2010spectral}.

\noindent Domain adversarial training has been another prevalent approach, wherein a model learns to extract domain-invariant features via adversarial objectives. This approach was first introduced by Ganin et al. (2016) and has been extended by various works to improve its robustness and effectiveness\cite{ganin2016domain, liu2018multi, chen2018adversarial, kumarage2023j}. The idea is to train a feature extractor that minimizes the classification loss while maximizing the domain classification loss, thus promoting domain-invariant feature learning.

\noindent Self-training methods, which involve generating pseudo-labels for target domain data and iteratively refining the model, have also shown efficacy in DA tasks\cite{ye2020towards}. These methods start by training a model on the source domain and then using the trained model to assign labels to the target domain data\cite{bhattacharjee2023conda}. The model is then retrained with these pseudo-labels, gradually adapting to the target domain.

\noindent Multi-source domain adaptation (MSDA) extends DA to multiple source domains, aiming to use data from several domains to improve performance on a target domain. Methods like pairwise-MMD and adversarial autoencoders have been employed to handle domain discrepancies in MSDA settings\cite{li2018extracting, wu2016multi}. Pairwise-MMD calculates the discrepancy between each pair of the source domain and target domain, while adversarial autoencoders learn a shared latent space that minimizes domain differences. However, a critical limitation of these approaches is their reliance on unlabeled target domain data during training, which is often unfeasible in practice.

\subsection{Domain Generalization}

Domain generalization (DG) addresses the challenge of training models that generalize well to unseen target domains without access to any target domain data during training. This is particularly significant when obtaining labeled or unlabeled data from the target domain is impractical. Approaches in DG often focus on learning domain-invariant representations from multiple source domains\cite{muandet2013domain, ahadian2024mnist,wang2021learning}. One popular strategy for DG is invariant risk minimization (IRM), which aims to learn representations that are invariant across different domains by capturing the intrinsic relationships between features and labels\cite{arjovsky2019invariant}. IRM enforces the optimal classifier to remain invariant across different domains, ensuring that the learned representations generalize well to new, unseen domains. Another approach is the use of domain adversarial training, extended to handle multiple source domains, to learn generalized feature spaces\cite{ganin2016domain}. This involves training a domain discriminator that distinguishes between different source domains while the feature extractor learns to confuse the discriminator, leading to domain-invariant features. 

% Prompt-based methods have also been explored, where prompts help unify classification tasks across different domains by reducing the discrepancy between domain-specific vocabulary distributions and feature representations\cite{jia2022prompt}. Prompt-based learning leverages pre-trained language models by conditioning them with task-specific prompts, which align the objectives of language modeling and downstream tasks. This helps in improving the robustness of the models to domain shifts.

\noindent The advent of Supervised Contrastive Learning with memory-saving mechanisms has shown promise in improving DG performance by enhancing the alignment of features within the same class across domains\cite{tan2022domain}. Supervised Contrastive Learning (SCL) explicitly pulls together representations of the same class while pushing apart those of different classes, thereby improving the discriminative power of the learned features. By incorporating a memory bank, the approach can handle larger sets of contrasting examples without excessive memory consumption, leading to better generalization.

\subsection{Contrastive Learning}

Contrastive learning has emerged as a powerful technique for self-supervised and supervised learning, wherein the model learns to distinguish between similar and dissimilar examples. In the context of self-supervised learning, contrastive methods such as SimCLR and MoCo maximize the agreement between different augmentations of the same instance while minimizing the agreement between different instances\cite{chen2020simple, he2020momentum}. These methods typically rely on large batch sizes to ensure a diverse set of negative examples, although techniques like memory banks have been proposed to mitigate the associated computational costs. SimCLR, for instance, uses stochastic data augmentations and learns representations by maximizing agreement between different augmented views of the same data point using a contrastive loss in the latent space\cite{chen2020simple}. MoCo introduces a dynamic dictionary with a queue and a moving-averaged encoder to maintain a large set of negative examples efficiently\cite{he2020momentum}.

\noindent Supervised Contrastive Learning (SCL) extends this idea to labeled data by treating examples from the same class as positive pairs and those from different classes as negative pairs\cite{khosla2020supervised}. This approach has been successfully applied to various tasks, including image classification and natural language understanding, by improving the discriminative power of learned representations\cite{gunel2021supervised}. Another approach such as SimCSE, a simple contrastive learning framework that incorporates annotated pairs called entailment pairs as positives and contradiction pairs as hard negatives, helped refine textual embeddings, demonstrating the adaptability of contrastive learning in a semantic setting\cite{gao2021simcse, shaeri2023semi, bhattacharjee2023conda}. These negative samples were refined both in definition and in usage through CLINE, a method for generating both negative adversarial and contrastive examples to improve robustness under adversarial attacking, and a user-specified method to generate varying degrees of ``hardness'' of negative examples, highlighting the method’s ability to enhance performance in downstream tasks across multiple modalities\cite{wang2021cline, robinson2020contrastive}. In SCL, the model leverages label information to explicitly cluster representations of the same class and separate those of different classes, which helps in learning more robust features.

\noindent In the context of DG, SCL can be particularly effective as it explicitly encourages the model to learn domain-invariant features by pulling together representations of the same class from different domains and pushing apart those of different classes\cite{tan2022domain}. This is achieved by using a combination of contrastive loss and cross-entropy loss, ensuring that the learned features are both discriminative and generalizable across domains. Additionally, the use of a memory bank allows for the storage and reuse of representations from previous batches, enhancing the effectiveness of contrastive learning without incurring high memory costs\cite{tan2022domain}.

% In the case of In-context Learning (ICL) method [40], [41], we introduce several concrete data examples, enabling the model to emulate these examples and generate standard, meaningful, and accurate data. Through iterative improvements in data generation quality, we observed that summarizing the shortcomings of previously generated data and explicitly addressing these in the prompt significantly enhanced the quality of data generated after prompt updates.

\subsection{Large Language Models}

LLMs\cite{achiam2023gpt, touvron2023llama, vicuna2023} have proved themselves quite remarkably versatile in various applications of text generation\cite{tan2024large, mehrban2023evaluating}. One such field is the ability to mimic human writing and efficiently generate deceptive fake news with minimal prompting\cite{sun2024exploring}. Recent studies have attempted to address LLM-generated disinformation further 
including artificially generating robust propaganda with which to train\cite{chen2023can}.

\noindent Pertaining to the new flood of LLM-generated disinformation, whether malicious or accidental, detection models\cite{wu2023survey} and LLM-attribution methods\cite{li2023survey, rosenfeld2024whose} have risen in retaliation. One recent detection strategy takes into account style-related features of text to combat style-based attacks\cite{wu2023fake}. This approach, although promising in regards to tackling currently prevalent prompting engineering attacks, is not as suited to address future prompt attacks without the need to retrain models with constantly improving tactics\cite{kumarage2023reliable}. In this work, we focus on introducing a method employing domain generalization over various prompting techniques to diminish the need to retrain the LLM detector. 

\section{Methodology}

\subsection{Problem Definition}
In the context of domain generalization for model attribution, we have labeled data from \( k \) source domains, where each domain represents a prompting method. Each source domain \( \{D_{s_i}\}_{i=1}^{k} \) is denoted as \(D_{s_i} = \{X_{s_i}, Y_{s_i}\}\), where \( X_{s_i} \) is LLM-generated disinformation and \( Y_{s_i} \) is the LLM that generated it. The goal is to create a model that can generalize across different domains, ensuring it can accurately identify the source of disinformation regardless of the prompting method.

\noindent During training, only source domains are available, and the labeled dataset from the target domain \( D_{t} = \{X_{t}, Y_{t}\} \) is reserved for evaluation. Unlike multi-source domain adaptation (MSDA), domain generalization does not require additional unlabeled target domain data during training \cite{wu2016sentiment, ding2019learning, zhao2018adversarial, ramponi2020neural,singhal2023domain}. Instead, the model is trained to obtain domain-invariant features from the source domains. Consequently, the target domain data \( D_{t} \) is used exclusively for evaluation, allowing the trained model to make predictions on unseen domains.

\subsection{Model Architecture}

To achieve this, we employ a Supervised Contrastive Learning (SCL) approach. This method not only increases the margin of decision boundaries but also ensures uniform distribution within each class, thus minimizing domain discrepancies.

\noindent We start by shuffling all the source domain data and dividing the combined dataset into mini-batches. This ensures a stable domain distribution in each mini-batch. For a sampled mini-batch of size \( N \), denoted as \( S = \{x_i, y_i\}_{i=1}^{N} \), \( x_i \) and \( y_i \) represent the input disinformation and the LLM label respectively.
We use a pre-trained language model (PLM) as the encoder and take the hidden state of the last layer's [CLS] token as the document representation, denoted as \( h \). Next, the representations \( h \) are then fed into a feed-forward neural network \(f\) with a tanh activation function and layer normalization to produce reduced-dimension features \( z = f(h)\). These features are subsequently used by a classifier \( g \) for downstream tasks, and the cross-entropy loss \( L_{CE} \) is computed as follows:

\begin{equation}
   L_{CE} = -\frac{1}{N} \sum_{i=1}^{N} y_i \log(g(z_i)) 
\end{equation}
\noindent where g is a fully connected classifier.

\noindent Since domain generalization involves training an algorithm using multiple source domains, the main challenge is to learn an optimal joint hypothesis for these domains. The objective is to separate data points by their labels, thereby minimizing domain discrepancy within each class in the feature embedding space. Supervised Contrastive Learning (SCL) explicitly increases the margin of decision boundaries and promotes a uniform distribution within each class while also reducing the distances between source domains for each class\cite{khosla2020supervised}.

\noindent This strategy aligns perfectly with our aim to develop a unified hypothesis across various source domains. It allows the model to maintain a more consistent distribution for each label class. Specifically, within a given mini-batch \(S\) of size \(N\), the Supervised Contrastive Loss \( L_{SCL} \) is computed as follows:

\begin{equation}
    L_{SCL} = -\sum_{z_i \in S} \frac{1}{|P(i)|} \sum_{z_p \in P(i)} \log \frac{\exp(z_i \cdot z_p / \tau)}{\sum_{z_a \in A(i)} \exp(z_i \cdot z_a / \tau)}
\end{equation}
\noindent where \( P(i) \) represents the set of positive examples for \( z_i \), \( A(i) \) is the combined set of positive examples (e.g., disinformation generated by the same LLM) and negative examples (e.g., disinformation generated by different LLMs), and \( \tau \) is a scaling hyper-parameter (temperature).
\begin{figure}[H]
    \centering
    \includegraphics[width=0.55\columnwidth]{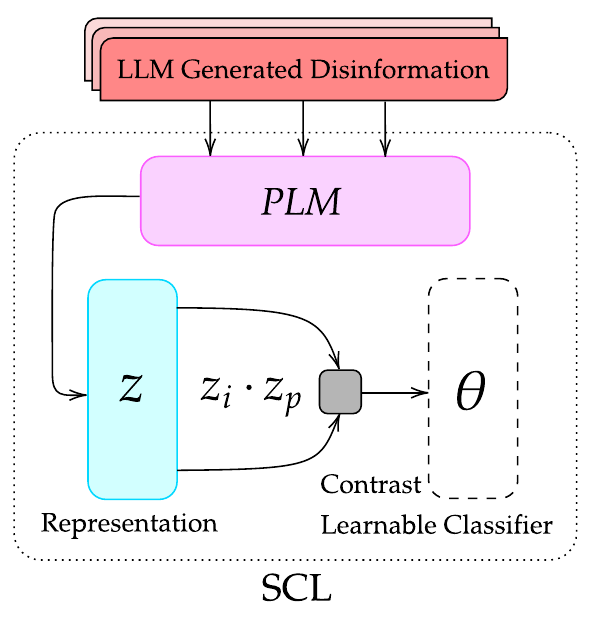}
    \caption{SCL Architecture.}
    \label{fig:SCL}
\end{figure}

\noindent Then, the combined loss function for training the model is:

\begin{equation}
    L = L_{CE} + L_{SCL}
\end{equation}

By employing this approach, the model can better align features of the same class and reduce domain discrepancies, leading to improved performance in identifying the source of disinformation across unseen domains. We provide the pseudo-code for our approach in Algorithm \ref{alg:supervised_contrastive_learning}

\begin{algorithm}
\caption{Algorithm for Supervised Contrastive Learning}
\label{alg:supervised_contrastive_learning}
\begin{algorithmic}[1]
\State \textbf{Input:} Batch size $N$, encoder $f$, classifier $g$
\For{$t \leq T_{\text{max}}$}
    \State Sample minibatch $S = \{x_i, y_i\}_{i=1}^{N}$
    \State $z = f(\text{PLM}(x))$
    \State $\mathcal{L}_{\text{CE}} = \frac{1}{N} \sum_{i=1}^{N} -y_i \log(g(z_i))$
    \State $P(i) \equiv \{z_j \in S, y_j = y_i\}$
    \State $A(i) \equiv \{z_j \in S, j \neq i\}$
    \State $\mathcal{L}_{\text{SCL}} = \sum_{z_i \in S} -\frac{1}{N} \sum_{z_p \in P(i)} \log \frac{\exp(z_i \cdot z_p / \tau)}{\sum_{z_a \in A(i)} \exp(z_i \cdot z_a / \tau)}$
    \State $\mathcal{L} = \mathcal{L}_{\text{CE}} + \mathcal{L}_{\text{SCL}}$
    \State update network by combined loss $\mathcal{L}$
\EndFor
\end{algorithmic}
\end{algorithm}

\section{Experiment}
\subsection{Dataset}
To test the performance of our model, we conducted experiments on the LLMFake\cite{chen2023can} dataset, a benchmark focused on disinformation generated by LLMs. LLMFake employs three disinformation generation methods: Hallucinated News, Arbitrary, and Controllable Generation. Hallucinated and Arbitrary methods prompt models like ChatGPT to produce fake news. Controllable Generation uses models like ChatGPT\footnote{\texttt{gpt-3.5-turbo}: \url{https://platform.openai.com/docs/models/gpt-3-5}}, Llama2\cite{touvron2023llama}, and Vicuna\cite{chiang2023vicuna} to create disinformation from nonfactual content in the Politifact, Gossipcop\cite{shu2020fakenewsnet}, and CoAID\cite{cui2020coaid} datasets through prompt techniques such as Paraphrasing, Rewriting, and Open-ended Generation. The Controllable Generation method results in approximately $5,400$ samples, equally distributed across three domains: Paraphrasing (P), Rewriting (R), and Open-ended Generation (O). From each domain, there are approximately $600$ LLM-generated disinformation samples generated by each of the three LLMs.

\subsection{Experiment Setting}
We conducted experiments in a cross-domain setting. The model is trained on one or two source domains and evaluated on the remaining two or one target domains, respectively (e.g., P, R-O or P-R, O). Model selection is based on validation performance on the combined test set of the source domains. For a fair comparison, we filter each domain based on semantic similarity, as the prompting methods are designed to preserve semantics and only change the style\cite{chen2023can}. After calculating the average semantic similarity, we retain only the disinformation samples with a similarity score above the average. Additionally, we split each domain into $600$ training examples, $200$ validation examples, and $200$ test examples. Since we do not have access to the target domain data, the training and model selection are based on the mixture distribution of source domains.

\noindent We use BERT\cite{devlin2018bert} and DeBERTa\cite{he2020deberta} as the default PLMs for encoders. For all of the encoders, we set the classifier representation dimension \(z\) to $256$. All models are trained with the Adam\cite{kingma2014adam} optimizer at a learning rate of \(1e^{-5}\). The whole model is trained up to $10$ epochs with a batch size of $16$ for training the BERT/DeBERTa baselines. All experiments are conducted on Nvidia 40GB V100 GPUs. We conduct a grid search for \(\tau \in \{0.1, 0.2, 0.5, 0.7, 0.8, 1.0\}\) in the P, R-O transfer (P and R as source domains, O as the target domain) and select \(\tau = 0.2\).

\subsection{Evaluation Metrics}
In this section, we discuss the evaluation metrics used to compare the performance of the models. Taking fine-tuned BERT-base as the baseline model, we calculate the actual difference in accuracy (\(act\_diff\)) and the improvement percentage (\(imp\)) from the target model to BERT-base for a given dataset as follows:

\begin{equation}
    act\_diff = acc\_tar - acc\_bert
\end{equation}

\begin{equation}
    imp = \frac{acc\_tar - acc\_bert}{acc\_bert} \times 100\%
\end{equation}
where \(acc\_tar\) indicates the accuracy of the target model and \(acc\_bert\) indicates the accuracy of the BERT-base model. We aggregate the actual differences and the improvement percentages across all test datasets to evaluate each model's performance. Additionally, we calculate the average accuracy across all test datasets for the target model. These results are included in the table.

\subsection{Comparison Models}
We compared our model to state-of-the-art models that were known to perform well on out-of-distribution datasets. The following models were used for comparison:

\noindent\textbf{Fine-tuned Language Models (LMs)} adapting pre-trained LMs to the model attribution task, and has proven effective in handling disinformation scenarios\cite{pelrine2021surprising}. We fine-tune BERT\cite{devlin2018bert} and DeBERTa\cite{he2020deberta} on the labeled data from source domains and directly test on the target domain. BERT is chosen for its robust performance in various NLP tasks and its ability to capture contextual information through bidirectional encoding\cite{devlin2018bert}, while DeBERTa is selected for its enhanced ability to handle long-range dependencies and improved mask decoding strategies\cite{he2020deberta}, making it effective in complex disinformation detection scenarios.

\noindent\textbf{Fine-tuning Setup} We fine-tune each pre-trained model on a dataset using cross-entropy loss. This approach is commonly used for model fine-tuning. To comprehensively evaluate the generalization ability of the pre-trained models, we conduct two types of experiments: 1) \textbf{full fine-tuning} and 2) \textbf{probing}. In the full fine-tuning experiment, we fine-tune each pre-trained language model (encoder) along with a classifier. While it is standard practice to fine-tune both the encoder and the classifier, this process often results in catastrophic forgetting \cite{mccloskey1989catastrophic}, which can obscure direct comparisons between pre-trained models. To address this, we also conduct probing experiments where we freeze the encoder (i.e., the pre-trained models) and train only the classifier with a single linear layer. This method allows for a more direct comparison of the pre-trained representations, similar to the approach in\cite{aghajanyan2020better}.

\subsection{Experiment Results}
We conducted extensive experiments to evaluate the performance of our proposed model, SCL\textsubscript{BERT}, compared to BERT and DeBERTa baselines using the FakeLLM dataset. The results presented in Table \ref{table:results} provide a comprehensive comparison of the performance in and out of the domain (OOD).

\begin{table*}[h]
\centering
\begin{tabular}{cccccccccccccc}
\toprule

\multicolumn{4}{c}{} & \multicolumn{5}{c}{Full} & \multicolumn{5}{c}{Probing} \\

\cmidrule(lr){5-9}\cmidrule(lr){10-14}

\multirow{2}{*}{Method} & \multicolumn{3}{c}{Domain} & \multirow{2}{*}{In Domain} & \multicolumn{3}{c}{Out of Domain} & \multirow{2}{*}{OOD Avg} & \multirow{2}{*}{In Domain} & \multicolumn{3}{c}{Out of Domain} & \multirow{2}{*}{OOD Avg}\\

\cmidrule(lr){2-4}\cmidrule(lr){6-8}\cmidrule(lr){11-13}

& \multicolumn{1}{c}{P} & \multicolumn{1}{c}{R} & \multicolumn{1}{c}{O} &  & \multicolumn{1}{c}{P} & \multicolumn{1}{c}{R} & \multicolumn{1}{c}{O} & & & \multicolumn{1}{c}{P} & \multicolumn{1}{c}{R} & \multicolumn{1}{c}{O} & \\

\midrule

\multirow{7}{*}{BERT} & $\times$ & $\times$ & \checkmark & 79.40 & 26.75 & 53.69 & - & 41.01 &  71.39 & 22.98 & 51.44 & - & 38.03 \\
& $\times$ & \checkmark & $\times$ & 72.05 & 51.42 & - & 56.47 & 54.20 & 68.15 & 49.37 & - & 55.13 & 52.11 \\
& \checkmark & $\times$ & $\times$ & 58.18 & - & 70.55 & 42.46 & 55.91 & 52.73 & - & 65.47 & 38.14 & 52.32 \\
& $\times$ & \checkmark & \checkmark & 75.44 & 54.76 & - & - & 54.76 & 71.12 & 51.59 & - & - & 51.59 \\
& \checkmark & $\times$ & \checkmark & 62.72 & - & 68.93 & - & 68.93 & 57.62 & - & 62.54 & - & 62.54 \\
& \checkmark & \checkmark & $\times$ & 65.70 & - & - & 54.14 & 54.14 & 59.83 & - & - & 51.32 & 51.32 \\
\midrule
Upper-bound & \checkmark & \checkmark & \checkmark & 69.93 & - & - & - & - & 61.82 & - & - & - & - \\
\midrule
\midrule

\multirow{7}{*}{DeBERTa} & $\times$ & $\times$ & \checkmark & 82.69 & 45.32 & 42.37 & - & 43.76 & 71.21 & 42.89 & 40.06 & - & 41.71 \\
& $\times$ & \checkmark & $\times$ & 76.87 & 18.70 & - & 29.29 & 24.53 & 69.44 & 22.53 & - & 31.72 & 27.12 \\
& \checkmark & $\times$ & $\times$ & 62.59 & - & 64.89 & 37.47 & 50.60 & 58.89 & - & 59.11 & 39.56 & 50.13 \\
& $\times$ & \checkmark & \checkmark & 79.81 & 18.70 & - & - & 18.70 & 72.83 & 33.57 & - & - & 33.57 \\
& \checkmark & $\times$ & \checkmark & 73.94 & - & 78.86 & - & 78.86 & 64.72 & - & 70.12 & - & 70.12 \\
& \checkmark & \checkmark & $\times$ & 75.73 & - & - & 38.11 & 38.11 & 69.23 & - & - & 42.91 & 42.91 \\
\midrule
Upper-bound & \checkmark & \checkmark & \checkmark & 75.87 & - & - & - & - & 67.73 & - & - & - & - \\
\midrule
\bottomrule
\midrule

\multirow{21}{*}{\textbf{SCL\textsubscript{BERT}}} & \multirow{3}{*}{$\times$} & \multirow{3}{*}{$\times$} & \multirow{3}{*}{\checkmark} & 75.26 & 38.85 & 55.34 & - & \textbf{47.11} & 67.91 & 33.12 & 54.23 & - & \textbf{44.05} \\
&  &  &  & -4.11 & +12.1 & +1.65 & - & +6.10 & -3.48 & +10.14 & +2.79 & - & +6.02 \\
&  &  &  & -5.17\% & +45.23\% & +3.07\% & - & \textbf{+14.87\%} & -4.87\% & +44.12\% & +5.42\% & - & \textbf{+15.82\%} \\
\cmidrule(lr){2-14}

& \multirow{3}{*}{$\times$} & \multirow{3}{*}{\checkmark} & \multirow{3}{*}{$\times$} & 69.89 & 54.52 & - & 58.34 & \textbf{56.81} & 66.32 & 55.12 & - & 56.89 & \textbf{55.34} \\
&  &  &  & -2.16 & +3.10 & - & +1.87 & +2.61 & -1.83 & +5.75 & - & +1.76 & +3.23 \\
&  &  &  & -2.99\% & +6.02\% & - & +3.31\% & \textbf{+4.81\%} & -2.68\% & +11.64\% & - & +3.19\% & \textbf{+6.19\%} \\
\cmidrule(lr){2-14}

& \multirow{3}{*}{\checkmark} & \multirow{3}{*}{$\times$} & \multirow{3}{*}{$\times$} & 56.34 & - & 71.25 & 46.71 & \textbf{56.83} & 50.13 & - & 68.83 & 45.12 & \textbf{53.82} \\
&  &  &  & -1.84 & - & +0.7 & +4.25 & +0.92 & -2.60 & - & +3.36 & +6.98 & +1.50 \\
&  &  &  & -3.16\% & - & +0.99\% & +10.00\% & \textbf{+1.64\%} & -4.93\% & - & +5.13\% & +18.30\% & \textbf{+2.86\%} \\
\cmidrule(lr){2-14}

& \multirow{3}{*}{$\times$} & \multirow{3}{*}{\checkmark} & \multirow{3}{*}{\checkmark} & 72.91 & 59.13 & - & - & \textbf{59.13} & 67.29 & 55.71 & - & - & \textbf{55.71} \\
&  &  &  & -2.53 & +4.37 & - & - & +4.37 & -3.83 & +4.12 & - & - & +4.12 \\
&  &  &  & -3.35\% & +7.98\% & - & - & \textbf{+7.98\%} & -5.38\% & +7.98\% & - & - & \textbf{+7.98\%} \\
\cmidrule(lr){2-14}

& \multirow{3}{*}{\checkmark} & \multirow{3}{*}{$\times$} & \multirow{3}{*}{\checkmark} & 61.45 & - & 70.78 & - & \textbf{70.78} & 54.91 & - & 68.23 & - & \textbf{68.23} \\
&  &  &  & -1.27 & - & +1.85 & - & +1.85 & -2.71 & - & +5.69 & - & +5.69 \\
&  &  &  & -2.02\% & - & +2.68\% & - & \textbf{+2.68\%} & -4.70\% & - & +9.09\% & - & \textbf{+9.09\%} \\
\cmidrule(lr){2-14}

& \multirow{3}{*}{\checkmark} & \multirow{3}{*}{\checkmark} & \multirow{3}{*}{$\times$} & \textbf{69.34} & - & - & 59.70 & \textbf{59.70} & \textbf{63.11} & - & - & 57.52 & \textbf{57.52} \\
&  &  &  & +3.64 & - & - & +5.56 & +5.56 & +3.28 & - & - & +6.20 & +6.20 \\
&  &  &  & \textbf{+5.54\%} & - & - & +10.26\% & \textbf{+10.26\%} & \textbf{+5.48\%} & - & - & +12.08\% & \textbf{+12.08\%} \\
\midrule
\multirow{3}{*}{Upper-bound} & \multirow{3}{*}{\checkmark} & \multirow{3}{*}{\checkmark} & \multirow{3}{*}{\checkmark} & \textbf{72.61} & - & - & - & - & \textbf{65.21} & - & - & - & - \\
&  &  &  & +2.68 & - & - & - & - & +3.39 & - & - & - & - \\
&  &  &  & \textbf{+3.83\%} & - & - & - & - & \textbf{+5.48\%} & - & - & - & - \\

\bottomrule
\end{tabular}
\caption{Result of the models trained on FakeLLM dataset. In the full fine-tuning setup (denoted as Full), we train both the encoder and the classifier. In the probing setup (denoted as Probing), we freeze the encoder and train the classifier. For the proposed model, each domain setting includes three rows: ACCURACY (FIRST ROW), ACTUAL DIFFERENCE (SECOND ROW), and IMPROVEMENT PERCENTAGE (THIRD ROW). These results are compared against the BERT baseline model. The reported results are averaged across 3 runs for each setting.}
\label{table:results}
\end{table*}

\noindent\textbf{In-Domain Performance.} In the full fine-tuning setting, BERT achieved accuracies of 58.18\% on domain P, 72.05\% on domain R, and 79.40\% on domain O. DeBERTa showed better performance with accuracies of 62.59\% on P, 76.87\% on R, and 82.69\% on O. DeBERTa's superior performance can be attributed to several advancements in its architecture and training methods. It employs a disentangled attention mechanism and relative positional encodings, enhancing its ability to understand context and word relationships. Additionally, DeBERTa is more parameter-efficient, achieving better results with fewer parameters. It also benefits from improved training techniques and larger, more recent pre-training datasets, allowing it to capture contextual information more effectively. These improvements collectively result in DeBERTa's superior accuracy compared to BERT. However, in SCL\textsubscript{BERT}, we observe a slight drop in in-domain performance during domain generalization, which is often a necessary trade-off for better generalization to unseen domains. This trade-off results from regularization, learning diverse and robust features, and balancing performance across multiple domains. Consequently, the model sacrifices some specialization to become more versatile and robust across different domains.

% In the probing setting, all models' performance slightly decreased. This drop occurs because freezing the layers of LMs during fine-tuning prevents the model from adapting its pre-trained features to the specific requirements of the new task. The fixed, general representations from pre-training may not capture all the necessary information for the new task. This restriction limits the model's ability to learn new patterns or adjust to new data distributions. As a result, the classifier layer alone cannot compensate for the lack of fine-tuning in the deeper layers, leading to lower accuracy.

\begin{figure*}[t!]
     \centering
     \begin{subfigure}[b]{0.45\textwidth}
         \centering
         \includegraphics[width=\textwidth]{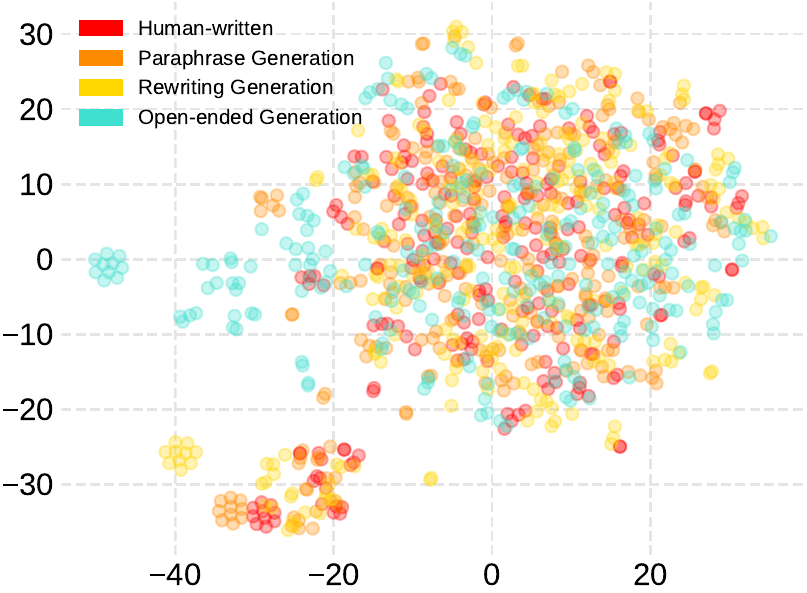}
         \caption{FakeLLM Dataset}
         \label{fig:vis_FakeLLM}
     \end{subfigure}
     \hfill
     \begin{subfigure}[b]{0.45\textwidth}
         \centering
         \includegraphics[width=\textwidth]{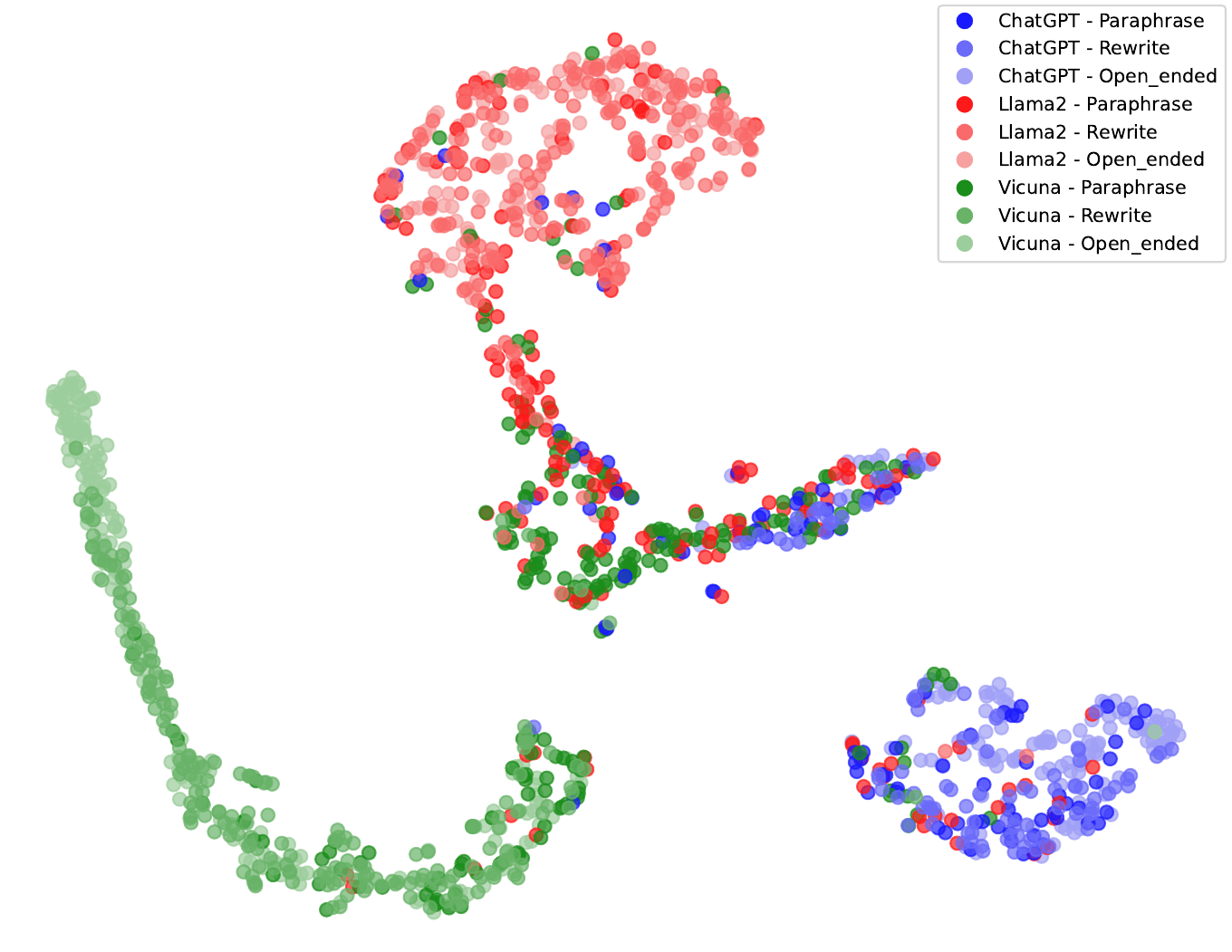}
         \caption{BERT}
         \label{fig:vis_T-SNE_BERT}
     \end{subfigure}
     \hfill
     \begin{subfigure}[b]{0.45\textwidth}
         \centering
         \includegraphics[width=\textwidth]{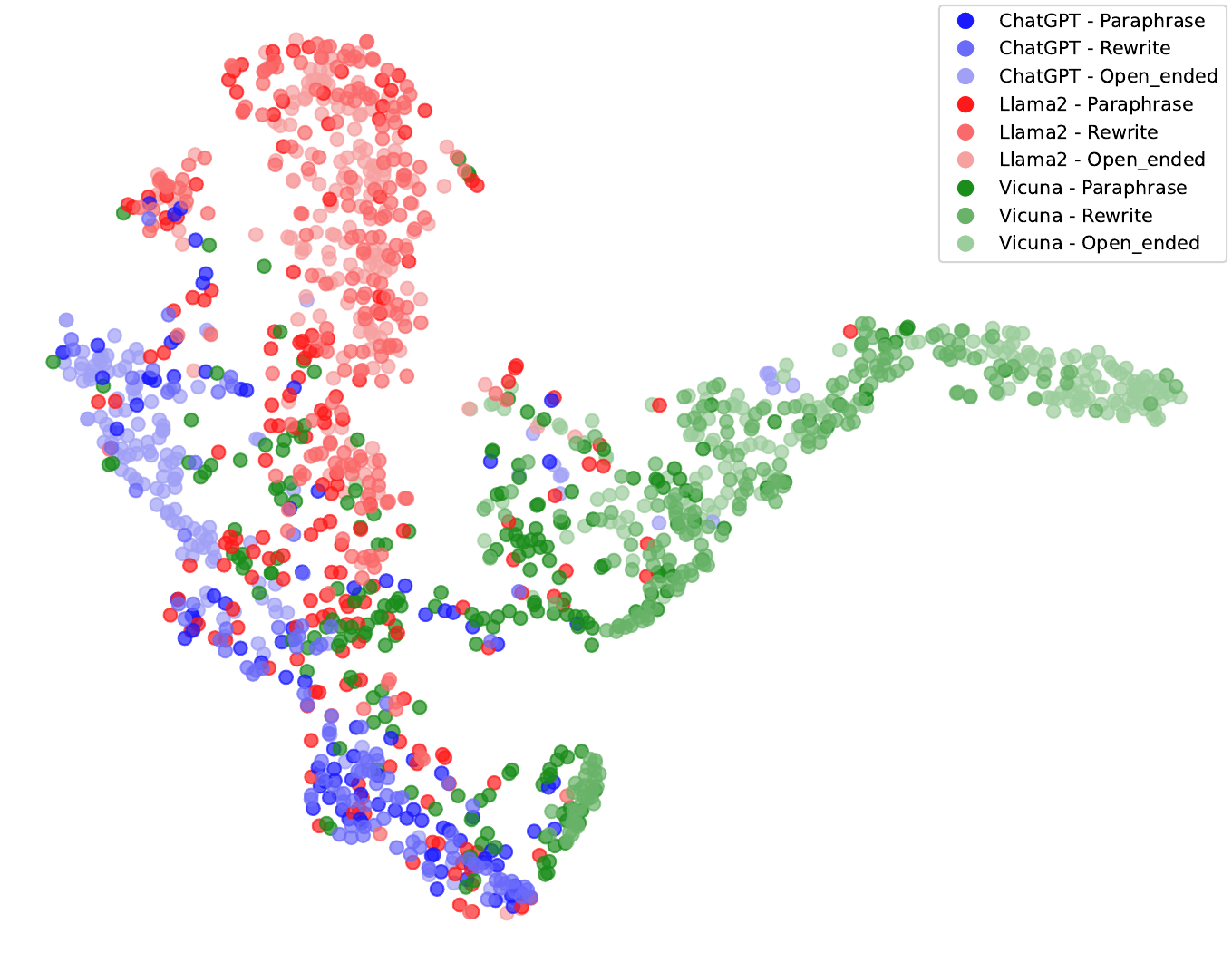}
         \caption{DeBERTa}
         \label{fig:vis_T-SNE_DeBERTa}
     \end{subfigure}
     \hfill
     \begin{subfigure}[b]{0.45\textwidth}
         \centering
         \includegraphics[width=\textwidth]{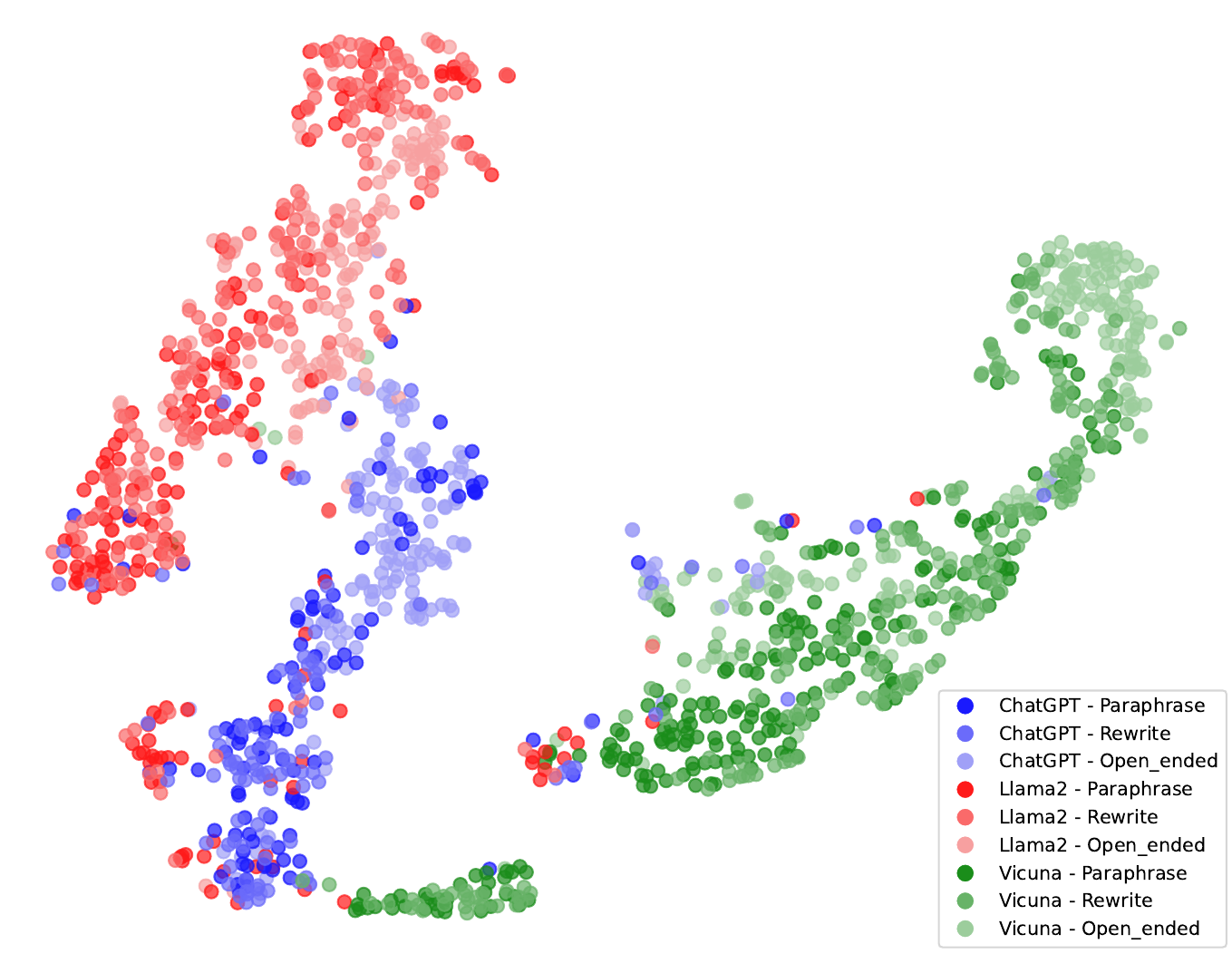}
         \caption{\textbf{SCL\textsubscript{BERT}}}
         \label{fig:vis_T-SNE_SCL}
     \end{subfigure}
     \hfill
     \caption{T-SNE visualization of 256-dimensional sentence embeddings $z$ for each model. The models are trained using O and R as source domains, and P as the target domain. For the visualization, 500 examples are sampled from each domain.}
     \label{fig:vis_T-SNE}
\end{figure*}

\noindent In the probing setting, all models’ performance slightly decreased. This drop occurs because freezing the layers of LMs during fine-tuning prevents the model from adapting its pre-trained features to the specific requirements of the new task. The fixed, general representations from pre-training may not capture all the necessary information for the new task. This restriction limits the model’s ability to learn new patterns or adjust to new data distributions. As a result, the classifier layer alone cannot compensate for the lack of fine-tuning in the deeper layers, leading to lower accuracy. While probing is intended to mitigate the risk of the classifier forgetting the learned representations by keeping the pre-trained weights fixed, it introduces a trade-off. On one hand, freezing the layers ensures that the rich, general-purpose features learned during pre-training are retained, preventing catastrophic forgetting. However, this same freezing restricts the model's adaptability, preventing it from refining its representations to better fit for the new task. This limitation can result in suboptimal performance since the model cannot fine-tune its pre-trained features to capture the specific patterns and information necessary for the new task. Therefore, while probing preserves the integrity of pre-trained representations, it limits the model's flexibility to adapt to new tasks. This dual effect leads to decreased performance as the model retains general knowledge but struggles to specialize, balancing between avoiding forgetting and achieving lower accuracy.\\

\noindent\textbf{Out-of-Domain Performance.}
In BERT and DeBERTa, the OOD performance dropped significantly. It shows the model struggles to adapt its pre-trained features to new, unseen domains. The PLMs may not capture the unique linguistic patterns and contexts of the out-of-domain data. Additionally, diverse and unseen patterns in OOD datasets reveal the models' difficulty in generalizing, as they may have overfitted to the training domain. This overfitting means they perform well on familiar data but poorly in different contexts. This highlights the importance of effective domain generalization to maintain robustness across various tasks and domains.

\noindent In both full fine-tuning and probing settings, SCL\textsubscript{BERT} shows substantial improvements in OOD performance. On average, performance improved by more than 7\% in full fine-tuning and 9\% in probing settings. This is because the clusters become more concentrated and domain discrepancies are significantly reduced. The performance boost indicates that increasing the margin of decision boundaries, promoting a uniform distribution within each class, and reducing distances between source domains by SCL\textsubscript{BERT} are effective strategies.

\noindent The experimental results clearly demonstrate that SCL\textsubscript{BERT} outperforms both BERT and DeBERTa baselines in the task of model attribution for LLM-generated disinformation. Notably, SCL\textsubscript{BERT} excels in out-of-domain scenarios, underscoring its robustness and generalizability across varied prompting methods and unseen datasets. These findings validate the efficacy of our proposed approach in enhancing the detection and attribution capabilities of models tasked with identifying the sources of LLM-generated disinformation. The consistent performance across both full fine-tuning and probing settings further highlights the strength of SCL\textsubscript{BERT} in maintaining robust and effective representations.

\begin{figure*}[t]
     \centering
     \begin{subfigure}[b]{0.49\textwidth}
         \centering
         \includegraphics[width=\textwidth]{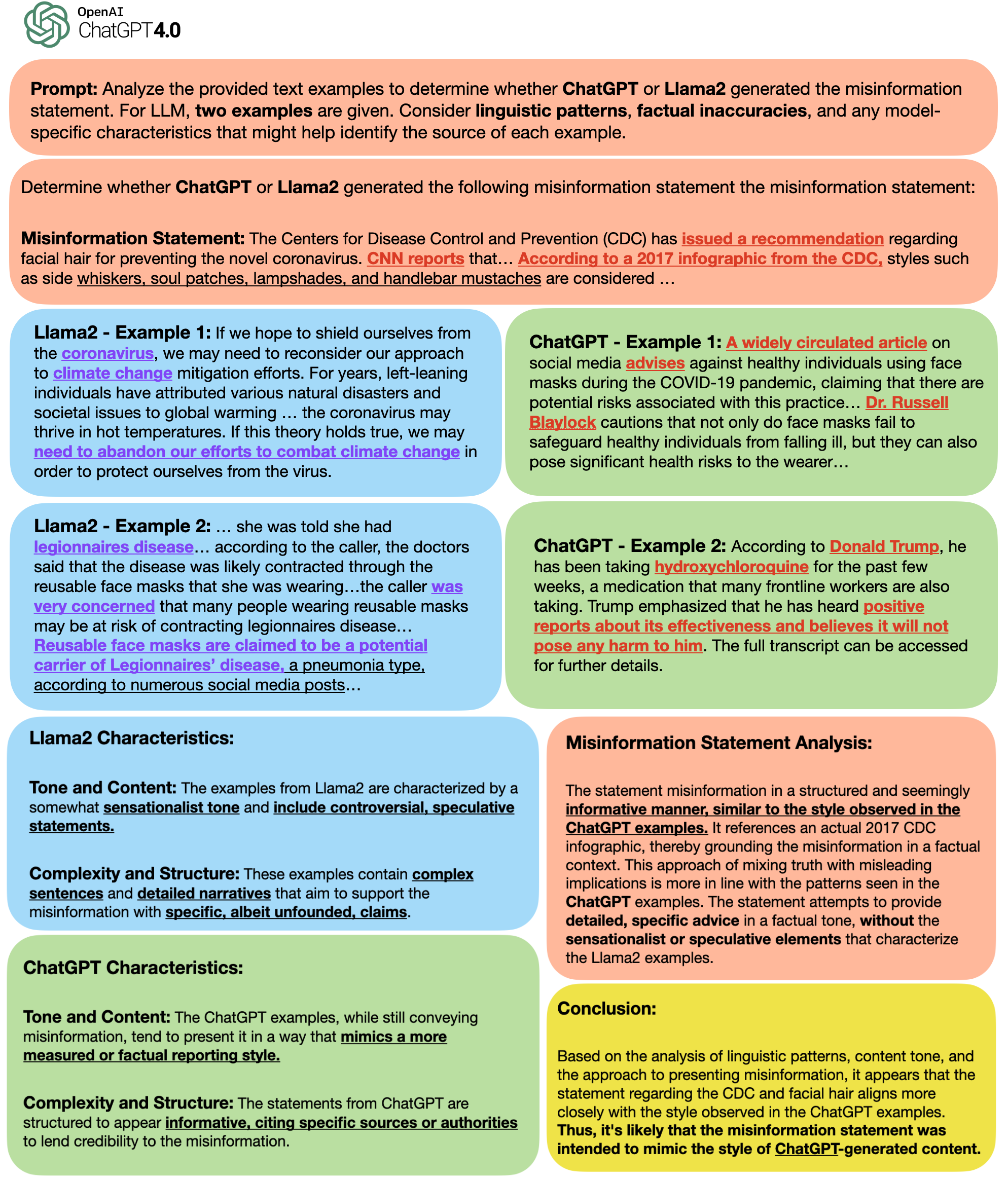}
         \caption{ICL Example 1 - ChatGPT4's Correct Detection}
         \label{fig:CaseStudyChatGPT4-Ex1}
     \end{subfigure}
     \hfill
     \begin{subfigure}[b]{0.49\textwidth}
         \centering
         \includegraphics[width=\textwidth]{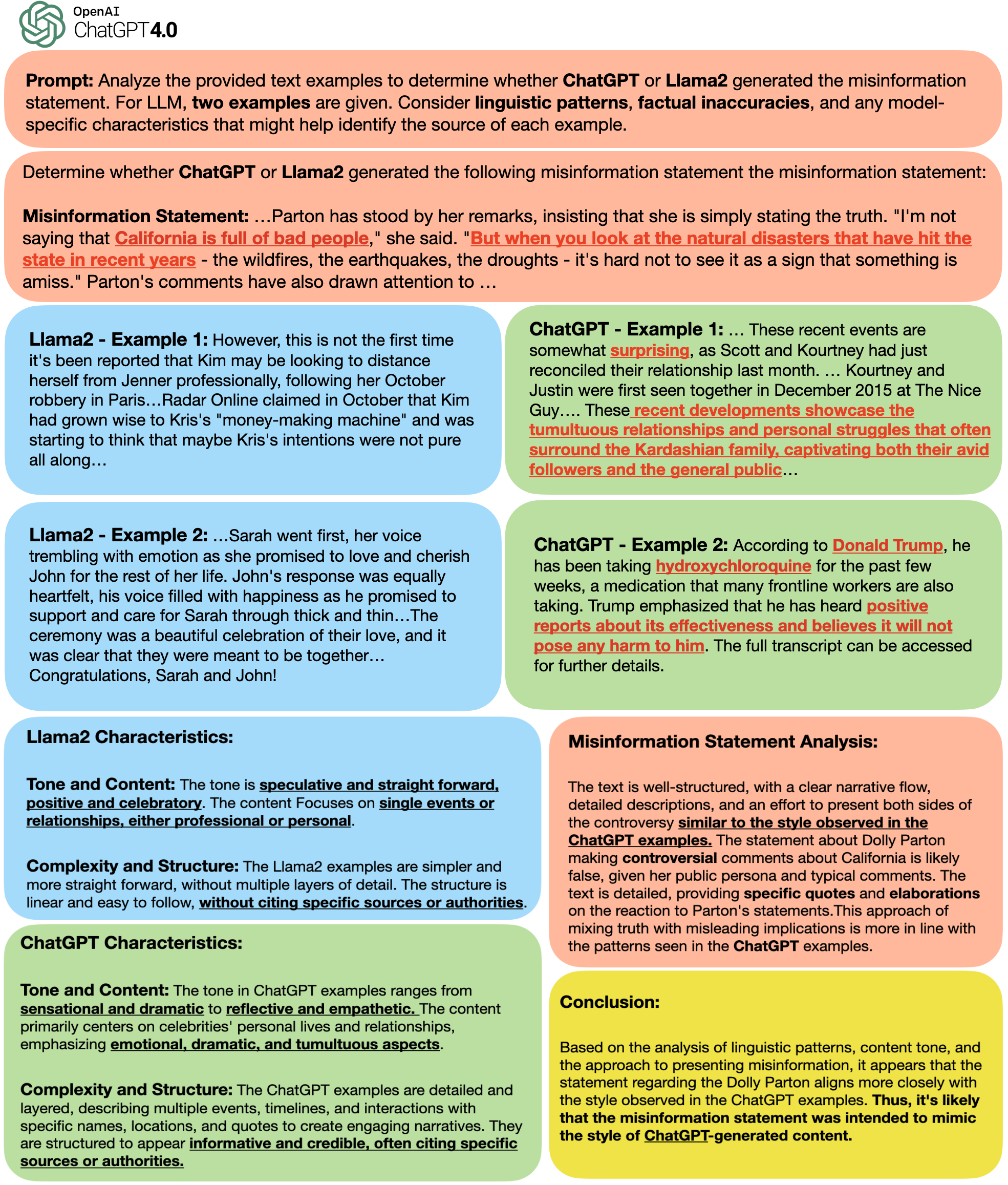}
         \caption{ICL Example 2 - ChatGPT4's Incorrect Detection}
         \label{fig:CaseStudyChatGPT4-Ex2}
     \end{subfigure}
     \hfill
     
     \caption{Investigating the Capacity of ChatGPT 4 for Attributing Disinformation Origin: An Analysis Using In-Context Learning Examples. This figure illustrates ChatGPT 4's potential in identifying the source of LLM-generated disinformation by analyzing the distinct writing styles of various language models through in-context learning. Examples showcase both successful and unsuccessful attributions, highlighting the method's reliance on the quality and representativeness of the input examples }
     \label{fig:CaseStudyChatGPT4}
\end{figure*}

\subsection{Analysis on Domain Divergence}

To further analyze our model's performance, we provide a detailed visualization of domain discrepancies using t-SNE\cite{van2008visualizing}. This technique helps us understand how our models handle variations across different domains. As illustrated in Figure \ref{fig:vis_FakeLLM}, disinformation generated by the three methods exhibits significant overlap in the latent space, particularly with human-written disinformation. This overlap suggests that these methods produce content that is difficult to distinguish from each other and from human-authored disinformation. 

\noindent In Figure \ref{fig:vis_T-SNE}, the models are trained with O and R as source domains and P as the target domain. For the visualization, we sampled $500$ examples from each domain. For the BERT model (Fig. \ref{fig:vis_T-SNE_BERT}), clear domain discrepancies are observed within each LLM cluster, and the clusters are relatively dispersed. This indicates that BERT struggles to produce domain-invariant representations. Similarly, the DeBERTa model (Fig. \ref{fig:vis_T-SNE_DeBERTa}) also shows domain discrepancies within the LLM clusters, demonstrating that DeBERTa does not significantly improve the ability to create domain-invariant features. In contrast, the SCL model (Fig. \ref{fig:vis_T-SNE_SCL}) presents more concentrated LLM clusters with significantly reduced domain discrepancies. This suggests that SCL\textsubscript{BERT} is more effective in producing domain-invariant representations, although some heterogeneity within the clusters remains. This visualization highlights the differences in performance between the models and their effectiveness in handling domain discrepancies. 
These visualizations confirm the superiority of SCL\textsubscript{BERT} in domain generalization and also underline the necessity of advanced techniques to improve model robustness across diverse domains.

\subsection{Case study}
In this section, we explore the capabilities of ChatGPT 4\cite{achiam2023gpt} in identifying the origins of LLM-generated disinformation through In-Context Learning (ICL). Specifically, we investigate whether ChatGPT 4 can effectively identify the source of disinformation when provided with example texts that illustrate the distinctive styles and patterns of LLMs.

\noindent ChatGPT 4 has shown potential in recognizing the writing styles of LLM-generated disinformation by utilizing in-context learning. In-context learning involves providing ChatGPT 4 with a set of examples that illustrate the specific styles and patterns of different LLMs. These examples serve as a reference for the model to understand and identify stylistic nuances (Fig. \ref{fig:CaseStudyChatGPT4}). During the study, we provided ChatGPT 4 with samples of disinformation generated by various LLMs, such as ChatGPT, Llama2, and Vicuna. The goal was to determine if ChatGPT 4 could detect the origin of these samples based on their writing style. ChatGPT 4 was able to correctly capture stylistic elements such as tone, word choice, sentence structure, and formatting in most cases(Fig. \ref{fig:CaseStudyChatGPT4-Ex1}). However, the success of this approach is heavily dependent on the quality and relevance of the provided examples. If the examples do not adequately represent the distinct styles of each LLM, ChatGPT 4's ability to accurately detect the origin of the disinformation diminishes\ref{fig:CaseStudyChatGPT4-Ex2}. Additionally, variations in the content and the inherent creativity of LLMs can introduce complexities that make reliable attribution challenging. Despite these challenges, the results of the study were promising. ChatGPT 4 demonstrated a notable ability to recognize and differentiate between the writing styles of different LLMs. Yet, this method alone proved insufficient for consistent and reliable LLM attribution. The accuracy of identifying the origin of the LLM-generated disinformation remains limited by the example quality and the subtle differences in writing styles.

\section{Conclusion}
In this paper, we proposed SCL\textsubscript{BERT}, a novel approach for model attribution in LLM-generated disinformation by treating it as a domain generalization problem. Our method leverages Supervised Contrastive Learning to reduce domain discrepancies and enhance decision boundaries. The experimental results demonstrated that SCL\textsubscript{BERT} outperforms existing baselines, including BERT and DeBERTa, particularly in out-of-domain scenarios. We observed that the application of SCL results in more concentrated clusters and significantly reduced domain discrepancies, with performance improvements of over 7\% in full fine-tuning and 9\% in probing settings.

\noindent We also conducted a case study using ChatGPT-4 to explore in-context learning capabilities for model attribution. The case study highlighted both the potential and limitations of using advanced language models for this task. While ChatGPT-4 demonstrated promising results, it also underscored the importance of providing high-quality and relevant training examples to maximize its effectiveness.

\noindent Despite these promising results, there are several directions for future work. First, we plan to explore the scalability of SCL\textsubscript{BERT} to larger and more diverse datasets to further validate its efficacy. Second, we aim to integrate advanced techniques to handle adversarial examples and improve robustness against noisy data~\cite{tao2023meta,tan2024glue}, common in real-world disinformation detection scenarios. Additionally, enhancing the interpretability of the model's decisions will provide deeper insights into the attribution process, contributing to the development of more transparent AI systems~\cite{tao2019fact,tan2024interpreting}. Finally, expanding the range of prompting methods and source LLMs in the training data could improve the model's generalizability, making it more resilient to the evolving landscape of disinformation generation.

\section*{Acknowledgments}

This work is supported by the U.S. Department of Homeland Security under Grant Award Number 17STQAC00001-08-00 and the National Science Foundation (NSF) under Grant Award Number IIS-2229461, SaTC-2241068, and IIS-2339198. The views and conclusions contained in this document are those of the authors and should not be interpreted as necessarily representing the official policies, either expressed or implied, of the U.S. Department of Homeland Security and the National Science Foundation.

\bibliographystyle{IEEEtran}
\bibliography{main}

\begin{thebibliography}{10}
\providecommand{\url}[1]{#1}
\csname url@samestyle\endcsname
\providecommand{\newblock}{\relax}
\providecommand{\bibinfo}[2]{#2}
\providecommand{\BIBentrySTDinterwordspacing}{\spaceskip=0pt\relax}
\providecommand{\BIBentryALTinterwordstretchfactor}{4}
\providecommand{\BIBentryALTinterwordspacing}{\spaceskip=\fontdimen2\font plus
\BIBentryALTinterwordstretchfactor\fontdimen3\font minus \fontdimen4\font\relax}
\providecommand{\BIBforeignlanguage}[2]{{%
\expandafter\ifx\csname l@#1\endcsname\relax
\typeout{** WARNING: IEEEtran.bst: No hyphenation pattern has been}%
\typeout{** loaded for the language `#1'. Using the pattern for}%
\typeout{** the default language instead.}%
\else
\language=\csname l@#1\endcsname
\fi
#2}}
\providecommand{\BIBdecl}{\relax}
\BIBdecl

\bibitem{chen2023can}
C.~Chen and K.~Shu, ``Can llm-generated misinformation be detected?'' \emph{arXiv preprint arXiv:2309.13788}, 2023.

\bibitem{jiang2024catching}
B.~Jiang, C.~Zhao, Z.~Tan, and H.~Liu, ``Catching chameleons: Detecting evolving disinformation generated using large language models,'' \emph{arXiv preprint arXiv:2406.17992}, 2024.

\bibitem{jiang2024media}
B.~Jiang, L.~Cheng, Z.~Tan, R.~Guo, and H.~Liu, ``Media bias matters: Understanding the impact of politically biased news on vaccine attitudes in social media,'' \emph{arXiv preprint arXiv:2403.04009}, 2024.

\bibitem{fallis2015disinformation}
D.~Fallis, ``What is disinformation?'' \emph{Library trends}, vol.~63, no.~3, pp. 401--426, 2015.

\bibitem{wu2019misinformation}
L.~Wu, F.~Morstatter, K.~M. Carley, and H.~Liu, ``Misinformation in social media: definition, manipulation, and detection,'' \emph{ACM SIGKDD explorations newsletter}, vol.~21, no.~2, pp. 80--90, 2019.

\bibitem{jiang2024disinformation}
B.~Jiang, Z.~Tan, A.~Nirmal, and H.~Liu, ``Disinformation detection: An evolving challenge in the age of llms,'' in \emph{Proceedings of the 2024 SIAM International Conference on Data Mining (SDM)}.\hskip 1em plus 0.5em minus 0.4em\relax SIAM, 2024, pp. 427--435.

\bibitem{huang2024can}
B.~Huang, C.~Chen, and K.~Shu, ``Can large language models identify authorship?'' \emph{arXiv preprint arXiv:2403.08213}, 2024.

\bibitem{kumarage2024survey}
T.~Kumarage, G.~Agrawal, P.~Sheth, R.~Moraffah, A.~Chadha, J.~Garland, and H.~Liu, ``A survey of ai-generated text forensic systems: Detection, attribution, and characterization,'' \emph{arXiv preprint arXiv:2403.01152}, 2024.

\bibitem{arbel2019maximum}
M.~Arbel, D.~J. Sutherland, M.~Bi{\'n}kowski, and A.~Gretton, ``Maximum mean discrepancy gradient flow,'' in \emph{Advances in Neural Information Processing Systems}, 2019, pp. 12\,054--12\,064.

\bibitem{pan2010spectral}
S.~J. Pan, I.~W.-H. Tsang, J.~T. Kwok, and Q.~Yang, ``Spectral feature alignment for unsupervised domain adaptation,'' \emph{arXiv preprint arXiv:0909.0507}, 2010.

\bibitem{borgwardt2006integrating}
K.~M. Borgwardt, A.~Gretton, M.~J. Rasch, H.-P. Kriegel, B.~Sch{\"o}lkopf, and A.~J. Smola, ``Integrating structured biological data by kernel maximum mean discrepancy,'' in \emph{International Conference on Intelligent Systems for Molecular Biology}, 2006, pp. 49--57.

\bibitem{ganin2016domain}
Y.~Ganin, E.~Ustinova, H.~Ajakan, P.~Germain, H.~Larochelle, F.~Laviolette, M.~Marchand, and V.~Lempitsky, ``Domain-adversarial training of neural networks,'' in \emph{Journal of Machine Learning Research}, 2016, pp. 2096--2030.

\bibitem{liu2018multi}
M.~Liu, H.~Xu, D.~Tao, and M.~Xu, ``Multi-task adversarial network for disentangled feature learning,'' in \emph{Proceedings of the IEEE Conference on Computer Vision and Pattern Recognition}, 2018, pp. 3743--3751.

\bibitem{chen2018adversarial}
X.~Chen, L.~Zhang, D.~Zhang, M.~Wu, Q.~Xie, M.~Sun, and J.~Xiao, ``Adversarial deep ensemble: Evasion attacks and defenses for malware detection,'' in \emph{Proceedings of the 27th International Joint Conference on Artificial Intelligence}, 2018, pp. 4318--4324.

\bibitem{kumarage2023j}
T.~Kumarage, A.~Bhattacharjee, D.~Padejski, K.~Roschke, D.~Gillmor, S.~Ruston, H.~Liu, and J.~Garland, ``J-guard: Journalism guided adversarially robust detection of ai-generated news,'' \emph{arXiv preprint arXiv:2309.03164}, 2023.

\bibitem{ye2020towards}
Q.~Ye, Z.~Zhang, H.~Dai, J.~Xu, and H.~Li, ``Towards better concept transfer in multi-domain sentiment classification,'' in \emph{Proceedings of the 58th Annual Meeting of the Association for Computational Linguistics}, 2020, pp. 1077--1085.

\bibitem{bhattacharjee2023conda}
A.~Bhattacharjee, T.~Kumarage, R.~Moraffah, and H.~Liu, ``Conda: Contrastive domain adaptation for ai-generated text detection,'' \emph{arXiv preprint arXiv:2309.03992}, 2023.

\bibitem{li2018extracting}
D.~Li, Y.~Yang, Y.-Z. Song, and T.~M. Hospedales, ``Extracting domain-invariant features with domain adversarial networks,'' in \emph{Proceedings of the European Conference on Computer Vision}, 2018, pp. 589--605.

\bibitem{wu2016multi}
Z.~Wu and M.~Huang, ``Multi-source domain adaptation for neural sentiment classification,'' in \emph{Proceedings of the 2016 Conference on Empirical Methods in Natural Language Processing}, 2016, pp. 2255--2264.

\bibitem{muandet2013domain}
K.~Muandet, D.~Balduzzi, and B.~Sch{\"o}lkopf, ``Domain generalization via invariant feature representation,'' \emph{International Conference on Machine Learning}, pp. 10--18, 2013.

\bibitem{ahadian2024mnist}
P.~Ahadian, Y.~Feng, K.~Kosko, R.~Ferdig, and Q.~Guan, ``Mnist-fraction: Enhancing math education with ai-driven fraction detection and analysis,'' in \emph{Proceedings of the 2024 ACM Southeast Conference}, 2024, pp. 284--290.

\bibitem{wang2021learning}
H.~Wang, Q.~Yao, D.~Ramanan, and L.-P. Morency, ``Learning robust representations by projecting superfluous dimensions to noise,'' in \emph{Proceedings of the 37th International Conference on Machine Learning}, 2021, pp. 10\,010--10\,020.

\bibitem{arjovsky2019invariant}
M.~Arjovsky, L.~Bottou, I.~Gulrajani, and D.~Lopez-Paz, ``Invariant risk minimization,'' \emph{arXiv preprint arXiv:1907.02893}, 2019.

\bibitem{tan2022domain}
Q.~Tan, R.~He, L.~Bing, and H.~T. Ng, ``Domain generalization for text classification with memory-based supervised contrastive learning,'' in \emph{Proceedings of the 29th International Conference on Computational Linguistics}, 2022, pp. 6916--6926.

\bibitem{chen2020simple}
T.~Chen, S.~Kornblith, M.~Norouzi, and G.~Hinton, ``A simple framework for contrastive learning of visual representations,'' \emph{arXiv preprint arXiv:2002.05709}, 2020.

\bibitem{he2020momentum}
K.~He, H.~Fan, Y.~Wu, S.~Xie, and R.~Girshick, ``Momentum contrast for unsupervised visual representation learning,'' in \emph{Proceedings of the IEEE/CVF Conference on Computer Vision and Pattern Recognition}, 2020, pp. 9729--9738.

\bibitem{khosla2020supervised}
P.~Khosla, P.~Teterwak, C.~Wang, A.~Sarna, Y.~Tian, P.~Isola, A.~Maschinot, C.~Liu, and D.~Krishnan, ``Supervised contrastive learning,'' \emph{Advances in neural information processing systems}, vol.~33, pp. 18\,661--18\,673, 2020.

\bibitem{gunel2021supervised}
B.~Gunel, J.~Du, A.~Conneau, and V.~Stoyanov, ``Supervised contrastive learning for pre-trained language model fine-tuning,'' in \emph{Proceedings of the 59th Annual Meeting of the Association for Computational Linguistics}, 2021, pp. 713--728.

\bibitem{gao2021simcse}
T.~Gao, X.~Yao, and D.~Chen, ``Simcse: Simple contrastive learning of sentence embeddings,'' \emph{arXiv preprint arXiv:2104.08821}, 2021.

\bibitem{shaeri2023semi}
P.~Shaeri and A.~Katanforoush, ``A semi-supervised fake news detection using sentiment encoding and lstm with self-attention,'' in \emph{2023 13th International Conference on Computer and Knowledge Engineering (ICCKE)}.\hskip 1em plus 0.5em minus 0.4em\relax IEEE, 2023, pp. 590--595.

\bibitem{wang2021cline}
D.~Wang, N.~Ding, P.~Li, and H.-T. Zheng, ``Cline: Contrastive learning with semantic negative examples for natural language understanding,'' \emph{arXiv preprint arXiv:2107.00440}, 2021.

\bibitem{robinson2020contrastive}
J.~Robinson, C.-Y. Chuang, S.~Sra, and S.~Jegelka, ``Contrastive learning with hard negative samples,'' \emph{arXiv preprint arXiv:2010.04592}, 2020.

\bibitem{achiam2023gpt}
J.~Achiam, S.~Adler, S.~Agarwal, L.~Ahmad, I.~Akkaya, F.~L. Aleman, D.~Almeida, J.~Altenschmidt, S.~Altman, S.~Anadkat \emph{et~al.}, ``Gpt-4 technical report,'' \emph{arXiv preprint arXiv:2303.08774}, 2023.

\bibitem{touvron2023llama}
H.~Touvron, L.~Martin, K.~Stone, P.~Albert, A.~Almahairi, Y.~Babaei, N.~Bashlykov, S.~Batra, P.~Bhargava, S.~Bhosale \emph{et~al.}, ``Llama 2: Open foundation and fine-tuned chat models,'' \emph{arXiv preprint arXiv:2307.09288}, 2023.

\bibitem{vicuna2023}
\BIBentryALTinterwordspacing
W.-L. Chiang, Z.~Li, Z.~Lin, Y.~Sheng, Z.~Wu, H.~Zhang, L.~Zheng, S.~Zhuang, Y.~Zhuang, J.~E. Gonzalez, I.~Stoica, and E.~P. Xing, ``Vicuna: An open-source chatbot impressing gpt-4 with 90\%* chatgpt quality,'' March 2023. [Online]. Available: \url{https://lmsys.org/blog/2023-03-30-vicuna/}
\BIBentrySTDinterwordspacing

\bibitem{tan2024large}
Z.~Tan, A.~Beigi, S.~Wang, R.~Guo, A.~Bhattacharjee, B.~Jiang, M.~Karami, J.~Li, L.~Cheng, and H.~Liu, ``Large language models for data annotation: A survey,'' \emph{arXiv preprint arXiv:2402.13446}, 2024.

\bibitem{mehrban2023evaluating}
A.~Mehrban and P.~Ahadian, ``evaluating bert and parsbert for analyzing persian advertisement data,'' \emph{arXiv preprint arXiv:2305.02426}, 2023.

\bibitem{sun2024exploring}
Y.~Sun, J.~He, L.~Cui, S.~Lei, and C.-T. Lu, ``Exploring the deceptive power of llm-generated fake news: A study of real-world detection challenges,'' \emph{arXiv preprint arXiv:2403.18249}, 2024.

\bibitem{wu2023survey}
J.~Wu, S.~Yang, R.~Zhan, Y.~Yuan, D.~F. Wong, and L.~S. Chao, ``A survey on llm-gernerated text detection: Necessity, methods, and future directions,'' \emph{arXiv preprint arXiv:2310.14724}, 2023.

\bibitem{li2023survey}
D.~Li, Z.~Sun, X.~Hu, Z.~Liu, Z.~Chen, B.~Hu, A.~Wu, and M.~Zhang, ``A survey of large language models attribution,'' \emph{arXiv preprint arXiv:2311.03731}, 2023.

\bibitem{rosenfeld2024whose}
A.~Rosenfeld and T.~Lazebnik, ``Whose llm is it anyway? linguistic comparison and llm attribution for gpt-3.5, gpt-4 and bard,'' \emph{arXiv preprint arXiv:2402.14533}, 2024.

\bibitem{wu2023fake}
J.~Wu and B.~Hooi, ``Fake news in sheep's clothing: Robust fake news detection against llm-empowered style attacks,'' \emph{arXiv preprint arXiv:2310.10830}, 2023.

\bibitem{kumarage2023reliable}
T.~Kumarage, P.~Sheth, R.~Moraffah, J.~Garland, and H.~Liu, ``How reliable are ai-generated-text detectors? an assessment framework using evasive soft prompts,'' \emph{arXiv preprint arXiv:2310.05095}, 2023.

\bibitem{wu2016sentiment}
F.~Wu and Y.~Huang, ``Sentiment domain adaptation with multiple sources,'' in \emph{Proceedings of the 54th Annual Meeting of the Association for Computational Linguistics (Volume 1: Long Papers)}, 2016, pp. 301--310.

\bibitem{ding2019learning}
X.~Ding, Q.~Shi, B.~Cai, T.~Liu, Y.~Zhao, and Q.~Ye, ``Learning multi-domain adversarial neural networks for text classification,'' \emph{IEEE Access}, vol.~7, pp. 40\,323--40\,332, 2019.

\bibitem{zhao2018adversarial}
H.~Zhao, S.~Zhang, G.~Wu, J.~M. Moura, J.~P. Costeira, and G.~J. Gordon, ``Adversarial multiple source domain adaptation,'' \emph{Advances in neural information processing systems}, vol.~31, 2018.

\bibitem{ramponi2020neural}
A.~Ramponi and B.~Plank, ``Neural unsupervised domain adaptation in nlp---a survey,'' \emph{arXiv preprint arXiv:2006.00632}, 2020.

\bibitem{singhal2023domain}
P.~Singhal, R.~Walambe, S.~Ramanna, and K.~Kotecha, ``Domain adaptation: challenges, methods, datasets, and applications,'' \emph{IEEE access}, vol.~11, pp. 6973--7020, 2023.

\bibitem{chiang2023vicuna}
W.-L. Chiang, Z.~Li, Z.~Lin, Y.~Sheng, Z.~Wu, H.~Zhang, L.~Zheng, S.~Zhuang, Y.~Zhuang, J.~E. Gonzalez \emph{et~al.}, ``Vicuna: An open-source chatbot impressing gpt-4 with 90\%* chatgpt quality,'' \emph{See https://vicuna. lmsys. org (accessed 14 April 2023)}, vol.~2, no.~3, p.~6, 2023.

\bibitem{shu2020fakenewsnet}
K.~Shu, D.~Mahudeswaran, S.~Wang, D.~Lee, and H.~Liu, ``Fakenewsnet: A data repository with news content, social context, and spatiotemporal information for studying fake news on social media,'' \emph{Big data}, vol.~8, no.~3, pp. 171--188, 2020.

\bibitem{cui2020coaid}
L.~Cui and D.~Lee, ``Coaid: Covid-19 healthcare misinformation dataset,'' \emph{arXiv preprint arXiv:2006.00885}, 2020.

\bibitem{devlin2018bert}
J.~Devlin, M.-W. Chang, K.~Lee, and K.~Toutanova, ``Bert: Pre-training of deep bidirectional transformers for language understanding,'' \emph{arXiv preprint arXiv:1810.04805}, 2018.

\bibitem{he2020deberta}
P.~He, X.~Liu, J.~Gao, and W.~Chen, ``Deberta: Decoding-enhanced bert with disentangled attention,'' \emph{arXiv preprint arXiv:2006.03654}, 2020.

\bibitem{kingma2014adam}
D.~P. Kingma and J.~Ba, ``Adam: A method for stochastic optimization,'' \emph{arXiv preprint arXiv:1412.6980}, 2014.

\bibitem{pelrine2021surprising}
K.~Pelrine, J.~Danovitch, and R.~Rabbany, ``The surprising performance of simple baselines for misinformation detection,'' in \emph{Proceedings of the Web Conference 2021}, 2021, pp. 3432--3441.

\bibitem{mccloskey1989catastrophic}
M.~McCloskey and N.~J. Cohen, ``Catastrophic interference in connectionist networks: The sequential learning problem,'' in \emph{Psychology of learning and motivation}.\hskip 1em plus 0.5em minus 0.4em\relax Elsevier, 1989, vol.~24, pp. 109--165.

\bibitem{aghajanyan2020better}
A.~Aghajanyan, A.~Shrivastava, A.~Gupta, N.~Goyal, L.~Zettlemoyer, and S.~Gupta, ``Better fine-tuning by reducing representational collapse,'' \emph{arXiv preprint arXiv:2008.03156}, 2020.

\bibitem{van2008visualizing}
L.~Van~der Maaten and G.~Hinton, ``Visualizing data using t-sne.'' \emph{Journal of machine learning research}, vol.~9, no.~11, 2008.

\bibitem{tao2023meta}
Y.~Tao, ``Meta learning enabled adversarial defense,'' in \emph{2023 IEEE International Conference on Sensors, Electronics and Computer Engineering (ICSECE)}.\hskip 1em plus 0.5em minus 0.4em\relax IEEE, 2023, pp. 1326--1330.

\bibitem{tan2024glue}
Z.~Tan, C.~Zhao, R.~Moraffah, Y.~Li, S.~Wang, J.~Li, T.~Chen, and H.~Liu, ``" glue pizza and eat rocks"--exploiting vulnerabilities in retrieval-augmented generative models,'' \emph{arXiv preprint arXiv:2406.19417}, 2024.

\bibitem{tao2019fact}
Y.~Tao, Y.~Jia, N.~Wang, and H.~Wang, ``The fact: Taming latent factor models for explainability with factorization trees,'' in \emph{Proceedings of the 42nd international ACM SIGIR conference on research and development in information retrieval}, 2019, pp. 295--304.

\bibitem{tan2024interpreting}
Z.~Tan, L.~Cheng, S.~Wang, B.~Yuan, J.~Li, and H.~Liu, ``Interpreting pretrained language models via concept bottlenecks,'' in \emph{Pacific-Asia Conference on Knowledge Discovery and Data Mining}.\hskip 1em plus 0.5em minus 0.4em\relax Springer, 2024, pp. 56--74.

\end{thebibliography}

\end{document}